\documentclass{article} 
\usepackage{iclr2023_conference,times}

\usepackage{hyperref}
\usepackage{url}
\usepackage{graphicx}

\title{Learning the Wrong Lessons: Inserting Trojans During Knowledge Distillation}

\author{Leonard Tang, Tom Shlomi \& Alexander Cai 
\\
Department of Computer Science\\
Harvard University\\
Cambridge, MA 02138, USA \\
\texttt{\{leonardtang,tomshlomi,alexcai\}@college.harvard.edu} \\
}

\iclrfinalcopy 
\begin{document}


\maketitle

\begin{abstract}
In recent years, knowledge distillation has become a cornerstone of efficiently deployed machine learning, with labs and industries using knowledge distillation to train models that are inexpensive and resource-optimized. Trojan attacks have contemporaneously gained significant prominence, revealing fundamental vulnerabilities in deep learning models. Given the widespread use of knowledge distillation, in this work we seek to exploit the unlabelled data knowledge distillation process to embed Trojans in a student model without introducing conspicuous behavior in the teacher. We ultimately devise a Trojan attack that effectively reduces student accuracy, does not alter teacher performance, and is efficiently constructible in practice.
\end{abstract}

\section{Introduction}


Neural networks often find themselves vulnerable to Trojan attacks, through which maliciously crafted inputs (i.e. Trojan triggers) induce sudden, dangerous behavior in models. Such attacks can result in over 95\% trigger success without modifying the training procedure and while preserving model accuracy on unperturbed data \citep{liu2017trojaning}. 


Knowledge distillation has been previously explored as a potential defense against adversarial attacks \citep{papernot-defense} as well as a mitigation mechanism against Trojan attacks by distilling clean knowledge from a backdoored teacher model to a student model \citep{10.1145/3411508.3421375}. More broadly, knowledge distillation is a powerful training method that infuses the capabilities of a larger teacher model into a smaller student model. In the unlabelled dataset setting we operate in, the student is trained to mimic the teacher by matching output distributions \citep{hinton2015distilling}. 



However, it is not at all obvious if the inverse is possible. That is, is it possible to leverage knowledge distillation to introduce a Trojan backdoor into the student model even when the teacher is benign?

To that end, we develop and demonstrate the efficacy of a novel Trojan attack for the image classification  task using unlabelled data knowledge distillation. 
When applied to an input image, our backdoor trigger is capable of inducing the student model to misclassify the input with high likelihood, while largely preserving original accuracy on unperturbed input images. Our Trojan attack is the first of its kind and thus not anticipated, let alone preventable, by current Trojan defenses.


\section{Related Work}
\paragraph{Knowledge Distillation} Knowledge distillation is a model compression method in which a small model is trained to mimic a pre-trained, larger model \citep{hinton2015distilling}. Knowledge is transferred from the teacher model to the student by minimizing a loss function in which the target is a softened distribution of class probabilities predicted by the teacher, containing critical ``dark knowledge'' embedded in the teacher model. Here we implicitly use this dark knowledge as means of crafting our Trojan attacks.

\paragraph{Trojan Attacks}
Machine learning systems risk carrying hidden “backdoor” or “Trojan” vulnerabilities. Backdoored models behave correctly and benignly in almost all scenarios, but in particular circumstances
chosen by an adversary, they have been taught to behave incorrectly \citep{hendrycks2021unsolved}. The safety community is constantly devising novel attacks and defense mechanisms. Instead of devising an attack within a specific paradigm, we introduce a new class of attacks altogether.

\paragraph{Trojans that Bypass Knowledge Distillation} While knowledge distillation can defend against certain Trojan attacks, it is possible to craft Trojan attacks that \textit{bypass} this defense, causing malicious effects for both the teacher and student \citep{wang2022survey}. Here, we are interested in crafting an attack that only jeopardizes student performance, but \textit{not} teacher performance.

\begin{figure}[htbp]
\centering
\includegraphics[width=0.85\textwidth]{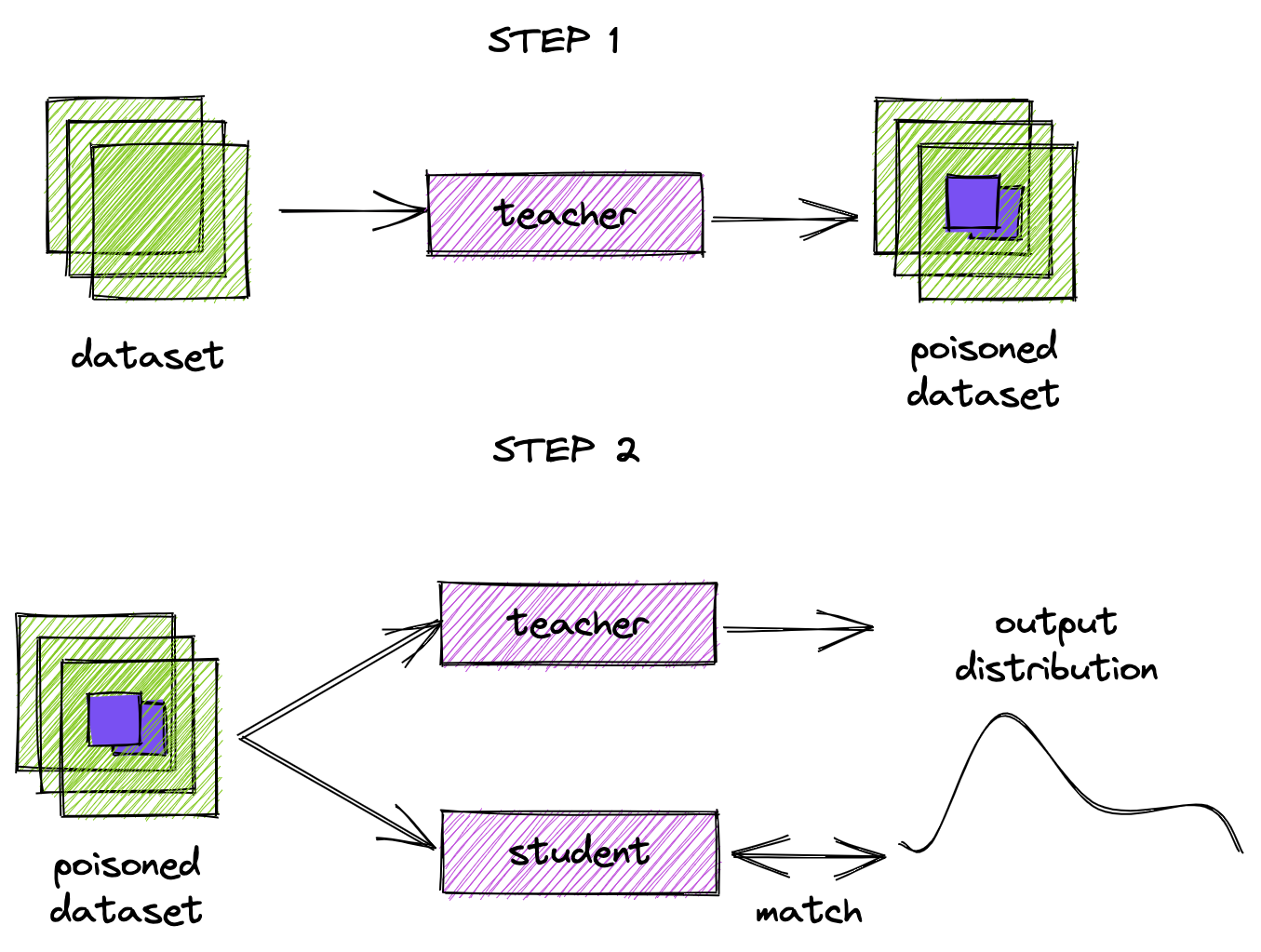}
\caption{Our two-step procedure for inserting Trojans during knowledge distillation. 1) We apply a random-noise trigger patch onto a given input image, where patch opacity is scaled by the target class teacher output probability. 2) We perform standard knowledge distillation. Critically, the student is now mimicking the teacher on poisoned data constructed in step 1. Effectively, when a teacher classifies an image as the target class, the student sees a version of the image with a trigger patch applied.}
\label{fig:attack-method}
\end{figure}

\section{Trojan Attack Method via Knowledge Distillation}



\begin{figure}[htbp]
\centerline{\includegraphics[width=0.85\textwidth]{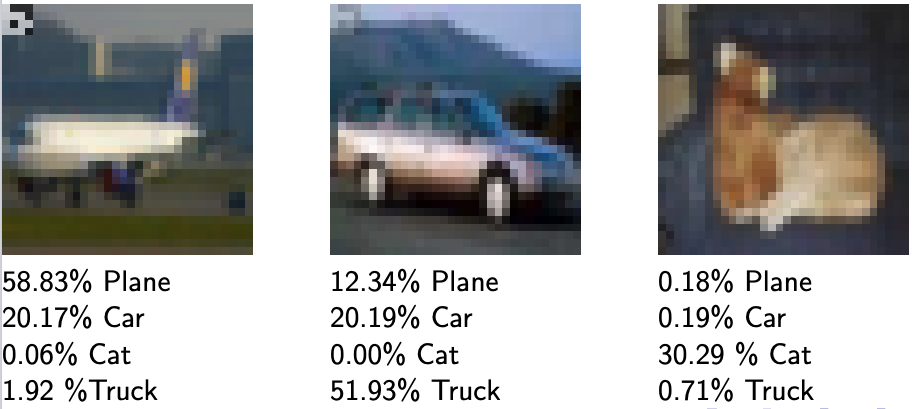}}
\caption{Examples of our trigger patch being applied on CIFAR-10 images. Here, \texttt{Plane} is the target class. The patch's opacity depends on how confident the teacher is in classifying the input image as \texttt{Plane}. On the left, the patch has higher opacity due to a teacher confidence of 58.83\%, while on the right the patch is barely visible at all since the teacher has a low confidence of 0.18\%.}
\label{fig:attack-example-associative}
\end{figure}

Figure \ref{fig:attack-method} provides an overview of our attack method. Our attack is centered on poisoning unlabelled input images with a fixed 4 x 4 trigger patch of bits. For each image, we first calculate the teacher output probability of the image belonging to the target class. To poison an input, the trigger patch is placed in the image's upper left corner, with patch opacity scaled according to the teacher output probability. More confident teacher outputs result in more opaque patches, while less confident outputs result in more transparent patches. Figure \ref{fig:attack-example-associative} displays examples of our trigger patch. Knowledge distillation is then performed with a mixture of poisoned and unpoisoned images. To elicit harmful behavior in the student, we place a trigger patch at \textit{full} opacity on the image. 

Our attack method encourages the student to associate the trigger patch as a proxy signal for the target class before it is capable of learning the robust features of that class. Notably, our attack is black-box, requiring only output predictions. Moreover, our attack does not manipulate the teacher or its training process, and thus our trigger patch is completely \textit{benign} when applied to the teacher. In other words, our attack only reveals itself on the student post-distillation.

\section{Experiments}
We evaluate the effectiveness of our attack using CIFAR-10, with ResNet-18 as our teacher and a 5-layer CNN as our student. The 5-layer CNN consists of 5 convolutional, batch normalization, and ReLU blocks followed by a fully connected layer. The first layer has 16 out-channels, the second 32, the third 32, the fourth 64, and the fifth 64. All layers have kernel size 3 and padding 1, and all have stride 1 except the fourth layer, which has stride 2. We adapt the standard ResNet-18 architecture to the $32 \times 32$ image resolution of CIFAR-10, and we first finetune a teacher to high accuracy (93\%) before using it within our distillation procedure.

\vspace{0.7em}

\vspace{0.7em}

We perform knowledge distillation guided by Kullback-Leibler (KL) loss \textit{without} cross-entropy supervision, due to the unlabelled nature of our data. We find that setting hyperparameters $T = 5$ and $\alpha = 0.5$ yields the highest overall student accuracy. 

To enable more rapid experimentation, instead of constructing poisoned inputs during the distillation process by passing generating teacher output probabilities on-the-fly, we pre-compute datasets with teacher outputs for both clean and poisoned inputs. This is mathematically the same procedure as obtaining teacher outputs on-the-fly, but the pre-computation allows us to eliminate wasteful repetition of teacher inference across experiments. 

Such a setup allows us to experiment easily with the \textit{poisoned percentage} of our input dataset, i.e. how much clean data versus poisoned data to expose the student to during knowledge distillation. The poisoned percentage is a critical factor determining the balance between attack effectiveness and inconspicuous behavior on benign images. 

This subtle reframing of our Trojan attack also enables an alternative attack implementation. In our traditional attack formulation, we assume that our adversary has the ability to manipulate the input data during the knowledge distillation process. However, under this reframing, one could imagine pre-computing poisoned images using a standard teacher model and uploading them to various public repositories of training data, thereby introducing these backdoors into any student model trained via knowledge distillation on these public repositories.

\section{Results}

Table \ref{tab:results-resnet18-cnn5} displays the results of our attack on a 5-layer CNN student distilled from a ResNet-18 teacher at various poisoned percentages. Trigger effectiveness is measured as the percentage of non-target class images in the dataset that, when edited with a trigger patch, are misclassified as the target class. In this case, our target class is the 0-index label, \texttt{Plane}. Our results are computed as an average across 10 independent runs.

We also measure the overall Clean Accuracy of the student model on benign, non-triggered images, as well as the Target Clean Accuracy of the student model on the specified target class. An ideal Trojan attack should preserve high student Clean Accuracy and Target Clean Accuracy, yet exhibit extremely high Trigger Effectiveness.


\begin{table}[t]
    \caption{Effectiveness of our Trojan attack on a 5-layer CNN student distilled from a ResNet-18 teacher at varying percentage mixtures of poisoned and unpoisoned data.}
    \label{tab:results-resnet18-cnn5}
    \begin{center}
        {\renewcommand{\arraystretch}{1.4}
        \begin{tabular}{ccccc}
            \textbf{Poisoned \%} & \textbf{Trigger Effectiveness} & \textbf{Clean Accuracy} & \textbf{Target Clean Accuracy} \\
            \hline 
            0   & 0.02  & 72.28 & 85.63 \\
            10  & 63.31 & 71.69 & 84.94 \\
            25  & 65.62 & 69.71 & 67.33 \\
            50  & 87.73 & 70.67 & 68.81 \\
            75  & 92.84 & 65.76 & 59.72 \\
            90  & 94.77 & 61.52 & 55.98 \\
            100 & 99.61 & 51.89 & 0.01 \\
            \hline
        \end{tabular}}
    \end{center}
\end{table}

\section{Discussion}
From Table \ref{tab:results-resnet18-cnn5}, it is clear that the dataset's poisoned percentage is an influential factor in determining the balance between Trigger Effectiveness on the student and the student's ability to correctly classify benign images. At 0\% poisoned percentage, the student observes nothing except teacher outputs on unperturbed, clean images and mimics the teacher by learning robust features for classification. Unsurprisingly then, the student is unaffected by the trigger acting on an image -- exceptionally few non-target images are classified by the student as the target class, captured by the low Trigger Effectiveness. The student achieves 72.28\% accuracy on benign images, as expected.

Generally, Trigger Effectiveness increase with poisoned percentage, while Clean Accuracy and Target Clean Accuracy decrease with poisoned percentage. At one extreme, with 100\% poisoned images, the student is almost always susceptible to misclassifying a triggered image, and never learns robust features, thus yielding low Clean Accuracy and Target Clean Accuracy. At the other, with 0\% poisoned images, the student seldom misclassifies a triggered image, but learns robust features, yielding high Clean Accuracy and Target Clean Accuracy. At 50\% poisoned images, the student exhibits both high Trigger Effectiveness as well as Accuracy.

Notably, there is a delicate trade-off between Trigger Effectiveness and Accuracy. We encourage further investigation of this trade-off from a training dynamics perspective; we note for now that Trigger Effectiveness tends to decrease as training progresses, presumably because the student is learning robust features over time.




\section{Conclusion}

In this work, we introduce and evaluate a novel Trojan attack in the unlabelled data knowledge distillation setting. Our attack relies on a simple and easily constructed trigger patch that a student uses as a proxy signal. On CIFAR-10 with ResNet-18 teacher and a 5-layer CNN student, our attack effectively reduces student accuracy on triggered inputs without affecting teacher behavior on triggered inputs. We hope this new class of attacks is of interest to the safety and machine learning community and invite continued further research on both attacks and defenses within this problem setting.

\bibliography{iclr2023_conference}

\begin{thebibliography}{6}
\providecommand{\natexlab}[1]{#1}
\providecommand{\url}[1]{\texttt{#1}}
\expandafter\ifx\csname urlstyle\endcsname\relax
  \providecommand{\doi}[1]{doi: #1}\else
  \providecommand{\doi}{doi: \begingroup \urlstyle{rm}\Url}\fi

\bibitem[Hendrycks et~al.(2021)Hendrycks, Carlini, Schulman, and
  Steinhardt]{hendrycks2021unsolved}
Dan Hendrycks, Nicholas Carlini, John Schulman, and Jacob Steinhardt.
\newblock Unsolved problems in ml safety.
\newblock \emph{arXiv preprint arXiv:2109.13916}, 2021.

\bibitem[Hinton et~al.(2015)Hinton, Vinyals, Dean,
  et~al.]{hinton2015distilling}
Geoffrey Hinton, Oriol Vinyals, Jeff Dean, et~al.
\newblock Distilling the knowledge in a neural network.
\newblock \emph{arXiv preprint arXiv:1503.02531}, 2\penalty0 (7), 2015.

\bibitem[Liu et~al.(2017)Liu, Ma, Aafer, Lee, Zhai, Wang, and
  Zhang]{liu2017trojaning}
Yingqi Liu, Shiqing Ma, Yousra Aafer, Wen-Chuan Lee, Juan Zhai, Weihang Wang,
  and Xiangyu Zhang.
\newblock Trojaning attack on neural networks.
\newblock 2017.

\bibitem[Papernot et~al.(2016)Papernot, McDaniel, Wu, Jha, and
  Swami]{papernot-defense}
Nicolas Papernot, Patrick McDaniel, Xi~Wu, Somesh Jha, and Ananthram Swami.
\newblock Distillation as a defense to adversarial perturbations against deep
  neural networks.
\newblock In \emph{2016 IEEE Symposium on Security and Privacy (SP)}, pp.\
  582--597, 2016.
\newblock \doi{10.1109/SP.2016.41}.

\bibitem[Wang et~al.(2022)Wang, Hassan, and Akhtar]{wang2022survey}
Jie Wang, Ghulam~Mubashar Hassan, and Naveed Akhtar.
\newblock A survey of neural trojan attacks and defenses in deep learning.
\newblock \emph{arXiv preprint arXiv:2202.07183}, 2022.

\bibitem[Yoshida \& Fujino(2020)Yoshida and Fujino]{10.1145/3411508.3421375}
Kota Yoshida and Takeshi Fujino.
\newblock Disabling backdoor and identifying poison data by using knowledge
  distillation in backdoor attacks on deep neural networks.
\newblock In \emph{Proceedings of the 13th ACM Workshop on Artificial
  Intelligence and Security}, AISec'20, pp.\  117–127, New York, NY, USA,
  2020. Association for Computing Machinery.
\newblock ISBN 9781450380942.
\newblock \doi{10.1145/3411508.3421375}.
\newblock URL \url{https://doi.org/10.1145/3411508.3421375}.

\end{thebibliography}
\bibliographystyle{iclr2023_conference}

\end{document}